\documentclass[letterpaper]{article}
\pdfoutput=1
\usepackage{aaai}
\usepackage{times}
\usepackage{helvet}
\usepackage{courier}
\usepackage{amsmath}
\usepackage{tikz}
\usepackage{numprint}
\usepackage{etoolbox,siunitx}
\usepackage{graphicx}
\robustify\bfseries
\usepackage{tcolorbox}

\usetikzlibrary{fit,positioning,arrows,automata,calc}
\tikzset{
  main/.style={circle, minimum size = 5mm, thick, draw=black!80, node distance = 10mm},
  connect/.style={-latex, thick},
  box/.style={rectangle, draw=black!100}
}
\begin{document}
% The file aaai.sty is the style file for AAAI Press 
% proceedings, working notes, and technical reports.
%
\title{Keyphrase Extraction from Scholarly Articles as Sequence Labeling using Contextualized Embeddings}
% \author{Paper ID - 9173}
\author{Dhruva Sahrawat\textsuperscript{\rm 3}\thanks{Contact for Dhruva Sahrawat: dhruva@comp.nus.edu.sg}, Debanjan Mahata\textsuperscript{\rm 1}\thanks{Contact for Debanjan Mahata: dmahata@bloomberg.net}, Raymond Zhang\textsuperscript{\rm 1}, Mayank Kulkarni\textsuperscript{\rm 1}, \\ \bf \Large Agniv Sharma \textsuperscript{\rm 2}, \bf \Large Rakesh Gosangi \textsuperscript{\rm 1}, \bf \Large Amanda Stent \textsuperscript{\rm 1}, \bf \Large Yaman Kumar \textsuperscript{\rm 2}, \\ \bf \Large Rajiv Ratn Shah \textsuperscript{\rm 2}, \bf \Large Roger Zimmermann \textsuperscript{\rm 3}  \\ % All authors must be in the same font size and format. Use \Large and \textbf to achieve this result when breaking a line
\textsuperscript{\rm 1}Bloomberg, USA \\ 
%If you have multiple authors and multiple affiliations
% use superscripts in text and roman font to identify them. For example, Sunil Issar,\textsuperscript{\rm 2} J. Scott Penberthy\textsuperscript{\rm 3} George Ferguson,\textsuperscript{\rm 4} Hans Guesgen\textsuperscript{\rm 5}. Note that the comma should be placed BEFORE the superscript for optimum readability
\textsuperscript{\rm 2} MIDAS-IIIT-Delhi, India \\
\textsuperscript{\rm 3} NUS, Singapore % email address must be in roman text type, not monospace or sans serif
}
\maketitle
\begin{abstract}
\begin{quote}
In this paper, we formulate keyphrase extraction from scholarly articles as a sequence labeling task solved using a BiLSTM-CRF, where the words in the input text are represented using deep contextualized embeddings. 
% AS - either change this to the ones presented in the experimental results, or add all 7 to the experimental results
We evaluate the proposed architecture using both contextualized and fixed word embedding models on three different benchmark datasets (Inspec, SemEval 2010, SemEval 2017), and compare with existing popular unsupervised and supervised techniques. Our results quantify the benefits of: (a) using contextualized embeddings (e.g. BERT) over fixed  word embeddings (e.g. Glove); (b) using a BiLSTM-CRF architecture with contextualized word embeddings over fine-tuning the contextualized word embedding model directly; and (c) using genre-specific contextualized embeddings (SciBERT). Through error analysis, we also provide some insights into why particular models work better than the others. Lastly, we present a case study where we analyze different self-attention layers of the two best models (BERT and SciBERT) to better understand the predictions made by each for the task of keyphrase extraction.
\end{quote}
\end{abstract}

\section{Introduction} \label{intro}
Keyphrase extraction is the process of selecting phrases that capture the most salient topics in a document \cite{turney2002learning}. Keyphrases serve as an important piece of document metadata, often used in downstream tasks including  information retrieval, document categorization, clustering and summarization. Keyphrase extraction has been applied to many different types of documents such as scientific papers \cite{kim2010semeval}, news articles \cite{hulth2006study}, Web pages \cite{yih2006finding}, and meeting transcripts \cite{liu2009unsupervised}.

Classic techniques for   keyphrase extraction involve a two stage approach \cite{hasan2014automatic}: (1) \textit{candidate generation}, and (2) \textit{pruning}. During the first stage, the document is processed to extract a set of candidate keyphrases. In the second stage, this candidate set is pruned by selecting the most salient candidate keyphrases, using either supervised or unsupervised techniques. In the supervised setting, pruning is formulated as a binary classification problem: \textit{determine if a given candidate is a keyphrase}. In the  unsupervised setting, pruning is treated as a \textit{ranking problem}, where the candidates are ranked based on some measure of \textit{importance} and those below a particular threshold are discarded. 

\noindent \textbf{Challenges} - Researchers typically employ a combination of different techniques for generating candidate keyphrases such as extracting named entities, finding noun phrases that adhere to  pre-defined lexical patterns \cite{Barker2000UsingNP}, or extracting n-grams that appear in an external knowledge base like Wikipedia \cite{Grineva:2009:EKT:1526709.1526798}. The candidates are further cleaned up using stop word lists or gazetteers. Errors in any of these techniques reduces the quality of candidate keyphrases. For example, if a named entity is not identified as such, it misses out on being considered as a keyphrase; if there are errors in part of speech tagging, extracted noun phrases might be incomplete. Also, since candidate generation involves a combination of heuristics with specific parameters, thresholds, and external resources, it is hard to reproduce any particular result or migrate implementations to new domains.

\noindent \textbf{Motivation} - Recently, researchers have started to approach keyphrase extraction as a sequence labeling task, where each token in the document is tagged as either being a part of a keyphrase or not \cite{gollapalli2016keyphrase,Alzaidy:2019:BSL:3308558.3313642}. There are many advantages to this new formulation. First, it completely bypasses the candidate generation stage and provides a unified approach to keyphrase extraction. 
% AS - I'm not sure sequence labeling is guaranteed to find an optimal assignment for long sequenes (across a whole document). Suggest to take this out.
Second, unlike binary classification where each keyphrase is classified independently, sequence labeling finds an optimal assignment of keyphrase labels for the entire document. Lastly, sequence labeling allows to capture long-term semantic dependencies in the document, which are known to be prevalent in natural language. 

More recently, there have been significant advances in deep contextual language models such as ELMo \cite{peters2018deep}, and BERT \cite{devlin2019bert}. These models can take an input text and provide contextual embeddings for each token for use in downstream architectures, or can perform task-specific fine-tuning. They have been shown to achieve state-of-the-art results for many different NLP tasks like document classification, question answering, dependency parsing, etc. More recent works \cite{Beltagy2019SciBERT,lee2019biobert} have shown that contextual embedding models trained on domain- or genre-specific corpora can outperform general purpose models that are usually trained on Wikipedia text.

\noindent \textbf{Contributions} - Despite all these developments, to the best of our knowledge, no  recent studies show the use of  contextual embeddings for keyphrase extraction. We expect that, as with other NLP tasks, keyphrase extraction can benefit from the use of contextual embeddings. We also posit that genre-specific language models may further help improve performance. To explore these hypotheses, in this paper, we approach keyphrase extraction as a sequence labeling task solved using a BiLSTM-CRF, where the underlying words in the input sequence are represented using various contextual embedding architectures. The following are the main contributions of this paper:

% \begin{itemize}
%     \item We study and quantify the benefits of using contextual embeddings in keyphrase extraction from scientific text when formulated as a sequence labeling task.  
%     \item We explore the benefits of using deep contextual language model (SciBERT) that is trained on a large corpus of scientific text and show the effectiveness of domain-specific embedding of words over generic embeddings (BERT) for the task of keyphrase extraction.
%     \item We perform our experiments on three benchmark datasets (Inspec, SemEval 2010, SemEval 2017), compare our models with popular baseline unsupervised and supervised techniques, and achieve state-of-the-art results for the task of keyphrase extraction.
%     \item We perform a thorough error analysis and provide insights into why particular contextual embeddings work better than the others, as well as present observations about the working of the different self-attention layers of the BERT based models, which achieve the top performances in our task.
% \end{itemize}

\noindent \textbf{$\bullet$} We quantify the benefits of using deep contextual embedding models (BERT, SciBERT, OpenAI GPT, ELMo, RoBERTa, Transformer XL and OpenAI GPT-2) for sequence-labeling-based keyphrase extraction from scientific text over using fixed word embedding models (word2vec, Glove and FastText). 

\noindent \textbf{$\bullet$} We demonstrate the benefits of using a BiLSTM-CRF architecture with contextualized word embeddings over fine-tuning the contextualized word embedding model to the keyphrase extraction task.  

\noindent \textbf{$\bullet$} We demonstrate improvements using contextualized word embeddings that are trained on a large corpus of in-genre text (SciBERT) over ones trained on generic text (BERT).

% AS - same comment about the seven deep LMs
\noindent \textbf{$\bullet$} We perform a robust set of experiments on three benchmark datasets (Inspec, SemEval-2010, SemEval-2017), and achieve state-of-the-art results. We compare the performance of these models with popular baseline unsupervised and supervised techniques.

% AS - transfer learning??
\noindent \textbf{$\bullet$} We perform a thorough error analysis and provide insights into how particular contextual embeddings work better than others
% show the effectiveness of transfer learning, as well as present observations about the 
and into the working of the different self-attention layers of our top-performing models.

\noindent \textbf{$\bullet$} We process the existing benchmark datasets using a B-I-O tagging scheme thus making them more suitable for building sequence-labeling-based models. We  make the processed datasets, our source code, and our trained models publicly available\footnote{www.anonymous.com} for the benefit of the research community.

The rest of the paper is organized as follows. Section \ref{background} presents prior work on keyphrase extraction and recent developments in deep contextualized language models. 
%Section \ref{problem} presents a formal definition of keyphrase extraction as sequence labeling. 
Section \ref{methods} details the BiLSTM-CRF architecture. Section \ref{experiments} describes our experiments. Sections \ref{Results} and \ref{sec:discussion} present experimental results and an attention analysis.

\section{Background and Related Work} \label{background}

\subsection{Keyphrase extraction} 
The task of automated keyphrase extraction has attracted attention from researchers for nearly 20 years \cite{Frank:1999:DKE:646307.687591}. Over this time, researchers have developed a wide array of both supervised and unsupervised techniques. In the supervised setting, keyphrase extraction is treated as a binary classification problem, with annotated keyphrases serving as positive examples and all other phrases as negative examples. Supervised techniques employ a machine learning model to determine if a given candidate phrase is a keyphrase based on  textual features such as term frequencies \cite{hulth2003improved}, syntactic properties \cite{kim2009re}, or location information \cite{10.1007/978-3-540-77094-7_41}. More features could be included with the use of external resources like document citations \cite{caragea2014citation} or hyperlinks \cite{Kelleher2005hypertext}. Different classification algorithms have been used, including naive bayes \cite{witten2005kea}, decision trees \cite{turney2000learning}, bagging \cite{hulth2003improved}, boosting \cite{hulth2001automatic}, neural networks \cite{lopez2010humb}, and SVMs \cite{zhao2011topical}.

Popular unsupervised methods such as TextRank \cite{mihalcea2004textrank}, LexRank \cite{erkan2004lexrank}, TopicRank \cite{bougouin2013topicrank}, SGRank \cite{danesh2015sgrank}, and SingleRank \cite{wan2008single}, directly leverage the graph-based ranking algorithm PageRank \cite{page1999pagerank}, with a combination of other heuristics based on tf-idf scores, word co-occurrence measures, extraction of specific lexical patterns, and clustering. Recently, several works \cite{mahata2018key2vec,wang2015using,mahata2018theme}, have shown the effectiveness of word embeddings in unsupervised keyphrase extraction. 

In the presence of domain-specific data, supervised methods have shown better performance.  Unsupervised methods have the advantage of not requiring any training data and can produce results in any domain. However, the assumptions of unsupervised methods do not hold for every type of document. On the other hand, casting keyphrase extraction as a binary classification problem has its disadvantages since each candidate phrase is labeled independent of all others. 

% Recasting keyphrase extraction as a classifica- tion problem has its weaknesses, however. Recall that the goal of keyphrase extraction is to identify the most representative phrases for a document. In other words, if a candidate phrase c1 is more representative than another candidate phrase c2, c1 should be preferred to c2. Note that a binary clas- sifier classifies each candidate keyphrase indepen- dently of the others, and consequently it does not allow us to determine which candidates are better than the others (Hulth, 2004; Wang and Li, 2011).

Gollapalli et al. \cite{gollapalli2016keyphrase} was one of the first works to  approach keyphrase extraction as a sequence labeling task. They used Conditional Random Fields (CRF) with many textual features such as tf-idf of the terms, orthographic information, POS tags, and positional information. Alzidy et al. \cite{Alzaidy:2019:BSL:3308558.3313642} used BiLSTM-CRFs, where the words were represented using fixed word embeddings like Glove embeddings \cite{pennington2014glove}. In this paper, we further explore sequence labeling techniques for keyphrase extraction using contextualized word embeddings. 

\subsection{Contextual Embeddings}
Recent research has shown that deep-learning language models trained on large corpora can significantly boost  performance on many NLP tasks and be effective in transfer learning \cite{peters2018deep,devlin2019bert,radford2018improving}. All these models aim to optimize for the traditional language model objective, which is to predict a word given its surrounding context. Therefore, they can assign each token a \textit{contextual} numerical representation that is a function of the entire input sentence. These contextual word embeddings are in contrast to the previously popular ``fixed" word embeddings that are learned using models such as Word2Vec \cite{goldberg2014word2vec}, Glove \cite{pennington2014glove}, and FastText \cite{joulin2016bag}. The word `apple' in the sentences - ``I bought apples from the farmers market", and ``I bought the new Apple IPhone", which is used in two different contexts, will be assigned the same numerical representation by a ``fixed" embedding model but a contextual model would differ the representation based on the sentence.  

ELMo \cite{peters2018deep} uses stacked bidirectional LSTMs with a residual connection to model sentences and the tokens are represented using character-level CNNs. BERT \cite{devlin2019bert} and OpenAI GPT \cite{radford2018improving} do away with the use of LSTMs and instead employ multi-layer Transformers \cite{vaswani2017attention} for sentence modeling. The main difference between BERT and OpenAI GPT is that the former uses a bidirectional transformer while the latter uses a left-to-right Transformer. OpenAI GPT-2 model \cite{radford2019language} was proposed as a direct successor of the GPT model, and is trained on 10x more data than the original GPT, with better performances and an additional benefit of working in a zero-shot transfer setting. RoBERTa \cite{liu2019roberta} was recently published as a replication study on BERT, where a new larger dataset was used for training, using more iterations, and removing the next sequence prediction training objective from the original model. This helped RoBERTa to achieve  state-of-the-art performance on different benchmark datasets making it equivalent in performance with other contextual language models like Transformer XL \cite{dai2019transformer}, and XLNet \cite{yang2019xlnet}. More recent works, like SciBERT \cite{Beltagy2019SciBERT}, have used the BERT architecture to build domain-specific language models. SciBERT was trained on a corpus of 1.14 million scientific papers mostly from computer science and biomedical domains. It has achieved state of the art results on many scientific NLP tasks including NER, document classification, and dependency parsing.

In this work, we combine the benefits of formulating keyphrase extraction as a sequence labeling task with the rich representation of language by contextual embeddings. To our knowledge, this is the first work which attempts a robust evaluation of the performance of a BiLSTM-CRF architecture using contextual embeddings for keyphrase extraction. We also present a thorough comparison of the performance of different word embedding models. 
% and BioBERT \cite{lee2019biobert}, have used the BERT architecture to build domain-specific language models. SciBERT was trained on a corpus of 1.14 million scientific papers mostly from computer science and biomedical domains. It has achieved state of the art results on many scientific NLP tasks including NER, document classificaiton, and dependency parsing. Likewise, BioERT was trained on only biomedical research articles from PubMed and PMC. Just like SciBERT, it attained state of the art results on biomedical NER, relation extraction, and question answering.

% %\input{related_work}

% %\input{problem_statement}

%\section{Problem Definition \label{problem}}
\section{Methodology}
\label{methods}
We approach the problem of automated keyphrase extraction from scholarly articles as a sequence labeling task, which can be formally stated as:
%\subsection{Problem definition}
\begin{tcolorbox}
\small Let $d = \{w_1, w_2, ..., w_n\}$ be an input text, where $w_t$ represents the $t\textsuperscript{th}$ token. Assign each $w_t$ in the document one of three class labels $Y=\{k_B, k_I, k_O\}$, where $k_B$ denotes that $w_t$ marks the beginning of a keyprahse, $k_I$ means that $w_t$ is inside a keyphrase, and $k_O$ indicates that $w_t$ is not part of a keyphrase.
\end{tcolorbox}

%\subsection{BiLSTM-CRF}
In this paper, we employ a BiLSTM-CRF architecture to solve this sequence labeling problem. LSTMs \cite{gers1999learning} are recurrent neural networks that deal with vanishing and exploding gradient problems with the use of gated architectures. Bidirectional LSTMs (BiLSTM) are generalization of LSTMs that capture long-distance dependencies between words in both directions.   

We first map each token $w_t$ in the input text to a fixed-size dense vector $x_t$, thus $d$ is represented as sequence of vectors $\boldsymbol{x} = \{x_1, x_2, ..., x_n\}$. The corresponding class labels are $\boldsymbol{y} = \{y_1, y_2, ..., y_n\}$, where ${y_i} \in Y$. We then use a BiLSTM to encode sequential relations between the word representations. A LSTM unit consists of four main components: input gate ($i_t$), forget gate ($f_t$), memory cell ($c_t$), and output gate ($o_t$), which are defined as below:
\begin{equation} \label{eq:1}
%\scriptstyle
i_t = \sigma(W_{xi}x_t + W_{hi}h_{t-1} + W_{ci}c_{t-1} + b_i)
\end{equation}
\begin{equation} \label{eq:2}
%\scriptstyle
f_t = \sigma(W_{xf}x_t + W_{hf}h_{t-1} + W_{cf}c_{t-1} + b_f)
\end{equation}
\begin{equation} \label{eq:3}
%\scriptstyle
c_t = f_t\odot c_{t-1} + i_t\odot\tanh(W_{xc}x_t + W_{hc}h_{t-1} + b_c)    
\end{equation}
\begin{equation} \label{eq:4}
%\scriptstyle
o_t = \sigma(W_{xo}x_t + W_{ho}h_{t−1} + W_{co}c_t + b_o)   
\end{equation}
\begin{equation} \label{eq:5}
%\scriptstyle
h_t = o_t\odot\tanh(c_t)
\end{equation}In the above equations, $\sigma$ denotes the sigmoid function, $\tanh$ the hyperbolic tangent function, and $\odot$ an  element-wise dot product. $W$ and $b$ are model parameters that are estimated during training, and $h_t$ is the hidden state. 

\begin{figure}[htbp]
    \centering
    \includegraphics[width=0.4\textwidth]{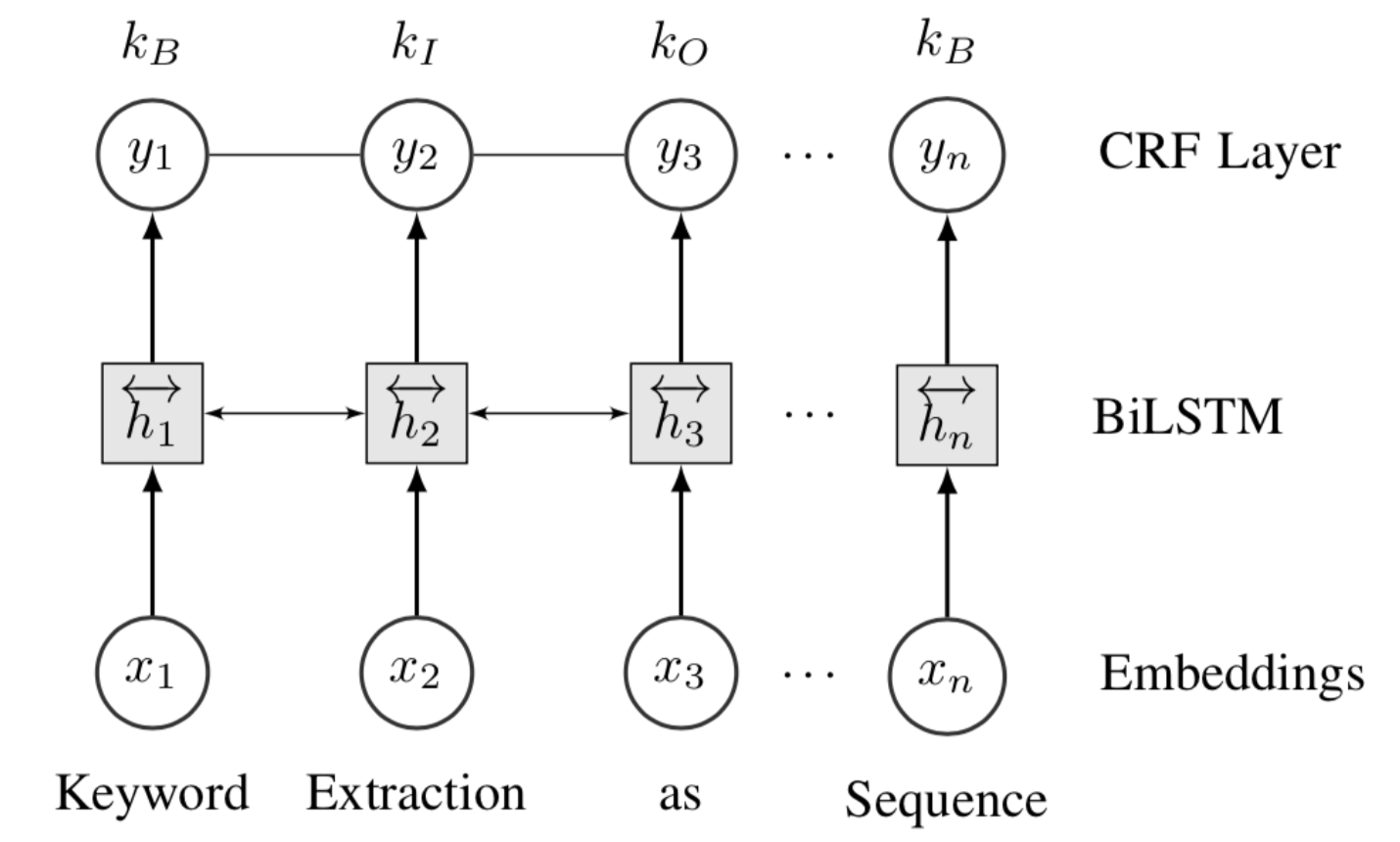}
    \caption{\small{BiLSTM-CRF architecture}}
    \label{fig:architecture}
\end{figure}

In a bidirectional LSTM, we apply equations \ref{eq:1} to \ref{eq:5} in both directions to create two hidden state vectors  $\overrightarrow{ht}$ and $\overleftarrow{ht}$, where $\overrightarrow{ht}$ provides a representation for word $w_t$ by incorporating information from the preceding words $\{w_1, ..., w_{t-1}\}$, and $\overleftarrow{ht}$ builds a representation for word $w_t$ by capturing information from the succeeding words $\{w_{t+1}, ..., w_n\}$. By concatenating $\overrightarrow{ht}$ and $\overleftarrow{ht}$, we get a vector representation of $w_t$ in the context of the input text $d$.
\begin{equation} \label{eq:6}
%\displaystyle
%\scriptstyle
\overleftrightarrow{h_t} = [\overrightarrow{h_t};\overleftarrow{h_t}]   
\end{equation} We then apply an affine transformation to map the output from the BiLSTM to the class space:
\begin{equation} \label{eq:7}
%\displaystyle
%\scriptstyle
f_t = W_a\overleftrightarrow{h_t}   
\end{equation} where $W_a$ is a matrix of size $|Y|\times2l$ and $l = |\overrightarrow{h_t}|$. 

The score outputs from the BiLSTM $\boldsymbol{f} = \{{f_1}, ..., {f_n}\}$ serve as input to a CRF layer. CRFs \cite{lafferty2001conditional} are discriminative probabilistic models that have been used in many sequence tagging problem in NLP. CRFs when used in conjunction with deep learning models \cite{DBLP:journals/corr/HuangXY15} have been shown to improve the performance of many sequence labeling tasks.

In a CRF, the score of an output label sequence is:
\begin{equation}
%\displaystyle
%\scriptstyle
s(\boldsymbol{f},\boldsymbol{y}) = \sum_{t=1}^{n} \tau_{y_{t-1},y_{t}} + f_{t,y_{t}}    
\end{equation}  $\tau$ is a transition matrix where $\tau_{i,j}$ represents the transition score from class $y_{t-1}$ to $y_t$. The likelihood for a labeling sequence is generated by exponentiating the scores and normalizing over all possible output label sequences.
\begin{equation}
%\displaystyle
%\scriptstyle
p(\boldsymbol{y}|\boldsymbol{f}) = \frac{\exp(s(\boldsymbol{f},\boldsymbol{y}))}{\sum_{y^{\prime}} \exp(s(\boldsymbol{f'},\boldsymbol{y'}))}    
\end{equation} During inference, CRFs use the Viterbi algorithm to efficiently find the optimal sequence of labels. The entire architecture is summarized in Figure \ref{fig:architecture}. As a baseline, we also experiment with a plain BiLSTM architecture.

\section{Experiments}
\label{experiments}
\noindent \textbf{Datasets} - We ran our experiments on three different publicly available keyphrase extraction datasets: Inspec \cite{hulth2003improved}, SemEval-2010 \cite{kim2010semeval} (hereafter, SE-2010), and SemEval-2017 \cite{augenstein2017semeval} (hereafter, SE-2017). Inspec consists of abstracts from 2000 scientific articles divided into train, validation and test sets containing 1000, 500, and 500 abstracts respectively. Each abstract is accompanied by two sets of human-annotated keyphrases: \textit{controlled} - as assigned by the authors, and \textit{uncontrolled} - assigned by the readers. Controlled keyphrases are mostly abstractive, i.e., not present in the abstracts, whereas the uncontrolled keyphrases are mostly extractive.  SE-2010  consists of 284 full length ACM articles divided into train, trial and test splits containing 144, 40, and 100 articles respectively.  SE-2010 also has author-assigned (controlled) and reader-assigned (uncontrolled) keyphrases.  SE-2017  consists of 500 open access articles published in ScienceDirect divided into train, dev and test sets containing 350, 50, and 100 articles respectively. Unlike the other two datasets,  SE-2017  provides location spans for all keyphrases, i.e. all keyphrases are extractive.

% AS why discarded the trial data from SemEval-2010?
Because we are modeling keyphrase extraction as a sequence labeling task, we only consider extractive keyphrases that are present in the article abstracts in each data set. For Inspec and SE-2010, we automatically identified the location spans for each extractive keyphrase. We discarded the full text articles from SE-2010 and SE-2017 due to memory constraints during training and inference with the contextual embedding models on large documents. We also discarded the trial dataset from SE-2010 and instead randomly chose 14 documents from the train split for validation. 

All the tokens in each dataset were tagged using the B-I-O tagging scheme described in the problem statement in the previous section. We plan to release this processed dataset along with this publication. Table \ref{tab:stats_table} provides some general statistics on the processed dataset used in this paper.

\begin{table*}[htbp]
\centering
\scalebox{1}{
\begin{tabular}{|c|c|c|c|c|c|c|ccc}
\hline
\textbf{Dataset} & \multicolumn{3}{c|}{\textbf{SE-2010}} & \multicolumn{3}{c|}{\textbf{Inspec}} & \multicolumn{3}{c|}{\textbf{SE-2017}} \\ \hline
 & \textbf{Train} & \textbf{Dev} & \textbf{Test} & \textbf{Train} & \textbf{Dev} & \textbf{Test} & \multicolumn{1}{c|}{\textbf{Train}} & \multicolumn{1}{c|}{\textbf{Dev}} & \multicolumn{1}{c|}{\textbf{Test}} \\ \hline
\textbf{\# Docs} & 130 & 14 & 100 & 1000 & 500 & 500 & \multicolumn{1}{c|}{350} & \multicolumn{1}{c|}{50} & \multicolumn{1}{c|}{100} \\ \hline
\textbf{\begin{tabular}[c]{@{}c@{}}Avg \# Keyphrases\end{tabular}} & 9.95 & 9.71 & 9.84 & 9.81 & 9.18 & 9.74 & \multicolumn{1}{c|}{15.62} & \multicolumn{1}{c|}{19.54} & \multicolumn{1}{c|}{17.69} \\ \hline
\textbf{\begin{tabular}[c]{@{}c@{}}Max Len of  Keyphrases\end{tabular}} & 6 & 6 & 6 & 8 & 8 & 9 & \multicolumn{1}{c|}{27} & \multicolumn{1}{c|}{19} & \multicolumn{1}{c|}{30} \\ \hline
\textbf{\begin{tabular}[c]{@{}c@{}}Avg  Len of Keyphrases\end{tabular}} & 1.72 & 1.84 & 1.73 & 2.15 & 2.15 & 2.15 & \multicolumn{1}{c|}{3.16} & \multicolumn{1}{c|}{2.72} & \multicolumn{1}{c|}{2.54} \\ \hline
% \textbf{\begin{tabular}[c]{@{}c@{}}\# B \\ Tokens\end{tabular}} & 1256 & 173 & 984 & 9790 & 4566 & 4844 & \multicolumn{1}{c|}{5469} & \multicolumn{1}{c|}{977} & \multicolumn{1}{c|}{1752} \\ \hline
% \textbf{\begin{tabular}[c]{@{}c@{}}\# I \\ Tokens\end{tabular}} & 959 & 87 & 716 & 11311 & 5263 & 5608 & \multicolumn{1}{c|}{11831} & \multicolumn{1}{c|}{1684} & \multicolumn{1}{c|}{2713} \\ \hline
% \textbf{\begin{tabular}[c]{@{}c@{}}\# O \\ Tokens\end{tabular}} & 21868 & 2559 & 18969 & 120407 & 56462 & 56848 & \multicolumn{1}{c|}{48020} & \multicolumn{1}{c|}{8597} & \multicolumn{1}{c|}{17566} \\ \hline
\textbf{\begin{tabular}[c]{@{}c@{}}Avg  \# Tokens\end{tabular}} & 185 & 201 & 207 & 142 & 133 & 135 & \multicolumn{1}{c|}{187} & \multicolumn{1}{c|}{225} & \multicolumn{1}{c|}{220} \\ \hline
\textbf{\begin{tabular}[c]{@{}c@{}}Max  \# Tokens\end{tabular}} & 432 & 416 & 395 & 557 & 330 & 384 & \multicolumn{1}{c|}{350} & \multicolumn{1}{c|}{399} & \multicolumn{1}{c|}{389} \\ \hline
\textbf{\begin{tabular}[c]{@{}c@{}}Min  \# Tokens\end{tabular}} & 55 & 111 & 65 & 15 & 16 & 23 & \multicolumn{1}{c|}{65} & \multicolumn{1}{c|}{137} & \multicolumn{1}{c|}{102} \\ \hline
\end{tabular}}
\caption{\small{General statistics of the processed dataset used in our experiments}}
\label{tab:stats_table}
\end{table*}

\noindent \textbf{Experimental Settings} - 
\label{experiment-settings}
One of the main aims of this work is to study the effectiveness of contextual embeddings in keyphrase extraction. To this end, we use the BiLSTM-CRF and BiLSTM architectures with seven different pre-trained contextual embeddings: BERT (small-cased, small-uncased, large-cased, large-uncased), SciBERT (basevocab-cased, basevocab-uncased, scivocab-cased, scivocab-uncased), OpenAI GPT, ELMo, RoBERTa (base, large), Transformer XL, and OpenAI GPT-2 (small, medium). As a baseline, we also use 300 dimensional fixed embeddings from Glove\footnote{https://nlp.stanford.edu/projects/glove/}, Word2Vec\footnote{https://github.com/mmihaltz/word2vec-GoogleNews-vectors}, and FastText\footnote{https://fasttext.cc/docs/en/english-vectors.html} (common-crawl, wiki-news). We also compare the proposed architecture against four popular baseline keyphrase extraction techniques: SGRank, SingleRank, Textrank, and KEA. Of these, the first three are unsupervised while KEA is a supervised technique. %Please refer Section \ref{background}, in order to get a brief overview of the different contextual and non-contextual embeddings as well as the baseline techniques used for the experiments.

We trained BiLSTM-CRF and BiLSTM models using Stochastic Gradient Descent (SGD) with Nesterov momentum in batched mode. Due to system memory constraints the batch size was set to 4. The learning rate was set to 0.05 and the models were trained for a total of 100 epochs with patience value of 4 and annealing factor of 0.5; i.e., if the model performance did not improve for 4 epochs, then the learning rate would be reduced by a factor of 0.5. The hidden layers in the BiLSTM models were set to 128 units and word dropout set to 0.05. The token representations obtained using different embeddings are not tuned during the training process. During inference, we run the model on a given abstract and identify keyphrases as all sequences of class labels that begin with the tag $k_B$ followed by zero or more tokens tagged $k_I$. As used by previous studies \cite{kim2010semeval}, we use Precision, Recall, and F1-measure based on actual matches against the ground-truth for evaluating the different approaches, and use the F1-measure to compare between different models\footnote{Due to space constraints we only report the F-measure in the paper.}.
%The supplementary material reports precision, recall and F-scores
%We assign the score of a keyphrase as the maximum confidence score obtained by that keyphrase across the document. We also performed experiments with taking the average of the confidence scores, but found that taking the maximum gives better performance. 

\section{Results}
\label{Results}
%We conducted a series of experiments to evaluate various components of our architecture. Following are some of the main findings from this work. 
In this section, we report the main observations for the different experiments that we performed. For each embedding model we  report results for the best performing variant of that model (e.g. cased vs uncased) on each dataset. Overall, BERT and SciBERT models are the best-performing embedding models, and the BiLSTM-CRF architecture is the best architecture, consistently across all the datasets.

\subsection{Architectures}

\noindent \textbf{CRF layer} - 
%Our first experiment aims to study the benefits of using CRF layer in the architecture. To this end, we trained two sets of models for all datasets and word embeddings. The first set of models use the BiLSTM-CRF and the other use only BiLSTM. 
Table \ref{tab:crf_table} presents a comparison of the BiLSTM and BiLSTM-CRF architectures in terms of F1-scores for three models: SciBERT, BERT, and ELMo. The addition of the CRF layer improved the performance for all datasets. However, it was most effective for ELMo. For example, on the SE-2010 data with ELMO, the CRF layer increased the F1-score by nearly 50\% from 0.157 to 0.225. An analysis of results on the SemEval-2017 data shows that the CRF layers is more effective in capturing keyphrases that include prepositions (e.g. `of'), conjunctions (e.g. `and'), and articles (e.g. `the'). We also observed that the CRF layer is more accurate with longer keyphrases (more than two tokens). 

\begin{table}[htbp]
\sisetup{table-column-width=16ex,round-mode=places,round-precision=3,detect-weight=true,detect-inline-weight=math}
    \centering
    \scalebox{1}{
    \begin{tabular}{|r|r|r|r|}
         \hline
          \multicolumn{4}{|c|}{SciBERT} \\ \hline
        & Inspec & SE-2010 & SE-2017  \\\hline
          BiLSTM-CRF & \textbf{0.593} & \textbf{0.357} & \textbf{0.521} \\
         BiLSTM     & 0.536 & 0.301 & 0.455 \\\hline 
          \multicolumn{4}{|c|}{BERT} \\ \hline
        & Inspec & SE-2010 & SE-2017  \\\hline
         BiLSTM-CRF & \textbf{0.591} & \textbf{0.330} & \textbf{0.522} \\
         BiLSTM      & 0.501 & 0.295   & 0.472 \\\hline
         \multicolumn{4}{|c|}{ELMo} \\ \hline
        & Inspec & SE-2010 & SE-2017  \\\hline
         BiLSTM-CRF & \textbf{0.568} & \textbf{0.225} & \textbf{0.504} \\
         BiLSTM      & 0.457 & 0.157   & 0.428 \\\hline
    \end{tabular}}
    \caption{\small{BiLSTM vs BiLSTM-CRF  (F1-score)}}
    \label{tab:crf_table}
\end{table}

\noindent \textbf{Fine-tuning} - 
Contextualized embedding models can be used in two ways: (1) they can serve as numerical representations of words that are to be used in downstream architectures, or (2) they can be fine-tuned to be optimized for a specific task. Fine-tuning typically involves adding an untrained layer at the end and then optimizing the layer weights  for the task-specific objective. We fine-tuned our best-performing contextualized embedding models (BERT and SciBERT) for each dataset and compared with the performance of the corresponding BiLSTM-CRF models when used with the same pre-trained embeddings. The results are summarized in Table \ref{tab:fine-tune}. 
The BiLSTM-CRF outperforms contextual embedding model fine-tuning across all datasets, for both BERT and SciBERT. We think this might be due to the small sizes of the datasets on which the models are fine-tuned.

\begin{table}[htbp]
\centering
\scalebox{1}{
\begin{tabular}{|c|c|c|c|}
\hline
\multicolumn{4}{|c|}{BERT} \\ \hline
 & \textbf{Inspec} & \textbf{SE-2010} & \textbf{SE-2017} \\ \hline
 \textbf{Fine-tuning} & 0.474 & 0.236 & 0.270 \\ \cline{2-4} 
 \textbf{BiLSTM-CRF} & \textbf{0.591} & \textbf{0.330} & \textbf{0.522} \\ \hline
\multicolumn{4}{|c|}{SciBERT} \\ \hline
 & \textbf{Inspec} & \textbf{SE-2010} & \textbf{SE-2017} \\ \hline
 \textbf{Fine-tuning} & 0.488 & 0.268 & 0.339 \\ \cline{2-4} 
 \textbf{BiLSTM-CRF} & \textbf{0.593} & \textbf{0.357} & \textbf{0.521} \\ \hline
\end{tabular}}
\caption{Fine-tuning vs Pretrained (F1-score)}
\label{tab:fine-tune}
\end{table}

\subsection{Contextual embeddings} 
Here we want to understand which of the various contextual embedding models that were discussed in Section \ref{background} is best suited for this task. These models vary in architecture, training data and hyperparameter choices. Table \ref{tab:contextual} presents the performance of the best variant (in our experiments) of the seven contextual embedding models and three fixed embedding models, using the BiLSTM-CRF architecture. 

Of the ten embedding architectures, BERT or BERT-based models consistently obtained the best performance across all datasets. This was expected considering that BERT uses bidirectional pre-training which is more powerful. SciBERT was consistently one of the top performing models and was significantly better than any of the other models on SemEval-2010. Further analysis of the results on SemEval-2010 shows that SciBERT was more accurate than other models in capturing keyphrases that contained scientific terms such as chemical names (e.g. `Magnesium', `Hydrozincite'), software projects (e.g. `HemeLB'), and abbreviations (e.g. `DSP', `SIMLIB'). SciBERT was also more accurate with keyphrases containing more than three tokens. The differences in training vocabulary is apparent where SciBERT is able to classify scientific nouns (e.g. `real time', `chip', `transform').

% \begin{table}[htbp]
%     \centering
%     \scalebox{0.8}{
%     \begin{tabular}{|l|r|r|r|r|r|}
%          \hline
%          & BERT & SBERT & ELMO & T-XL & OAI-GPT2 \\
%          \hline
%          Inspec & 0.591 & \textbf{0.593} & 0.567 & 0.511 & 0.531  \\
%          SemEval2010 & 0.330 & \textbf{0.357} & 0.224 & 0.221 & 0.239 \\
%          SemEval2017 & \textbf{0.521} & \textbf{0.521} & 0.504 & 0.445 & 0.438 \\
%          \hline
%     \end{tabular}}
%     \caption{\small{Comparison of various contextual embedding models.}}
%     \label{tab:contextual}
% \end{table}

\begin{table}[htbp]
    \centering
    \scalebox{1}{
    \begin{tabular}{|l|r|r|r|}
         \hline
         & Inspec & SE-2010 & SE-2017 \\
         \hline
         SciBERT & 0.593 & \textbf{0.357} & 0.521 \\
         BERT & 0.591 & 0.330 & \textbf{0.522} \\
         ELMo & 0.568 & 0.225 & 0.504 \\
         Transformer-XL & 0.521 & 0.222 & 0.445 \\
         OpenAI-GPT & 0.523 & 0.235 & 0.439 \\
         OpenAI-GPT2 & 0.531 & 0.240 & 0.439 \\
         RoBERTa & \textbf{0.595} & 0.278 & 0.508 \\
         Glove & 0.457 & 0.111 & 0.345 \\
         FastText & 0.524 & 0.225 & 0.426 \\
         Word2Vec & 0.473 & 0.208 & 0.292 \\
         \hline
    \end{tabular}}
    \caption{\small{Embedding models comparison (F1-score)}}
    \label{tab:contextual}
\end{table}

%SciBERT consistently did better than BERT and difference is most significant on SemEval2010. This is consistent our earlier hypothesis that language models trained on domain specific corpora can benefit keyphrase extraction.
%We expected OpenAI-GPT2 to outperform ELMo but our results demonstrate otherwise.  

Contextual embeddings outperformed their fixed counterparts for most of the experimental scenarios. The only exception was on SemEval-2010 where FastText outperformed Transformer-XL. Of the three fixed embedding models studied in this paper, FastText obtained the best performance across all datasets. 

\begin{figure}[htbp]
    \centering
    \includegraphics[width=0.35\textwidth]{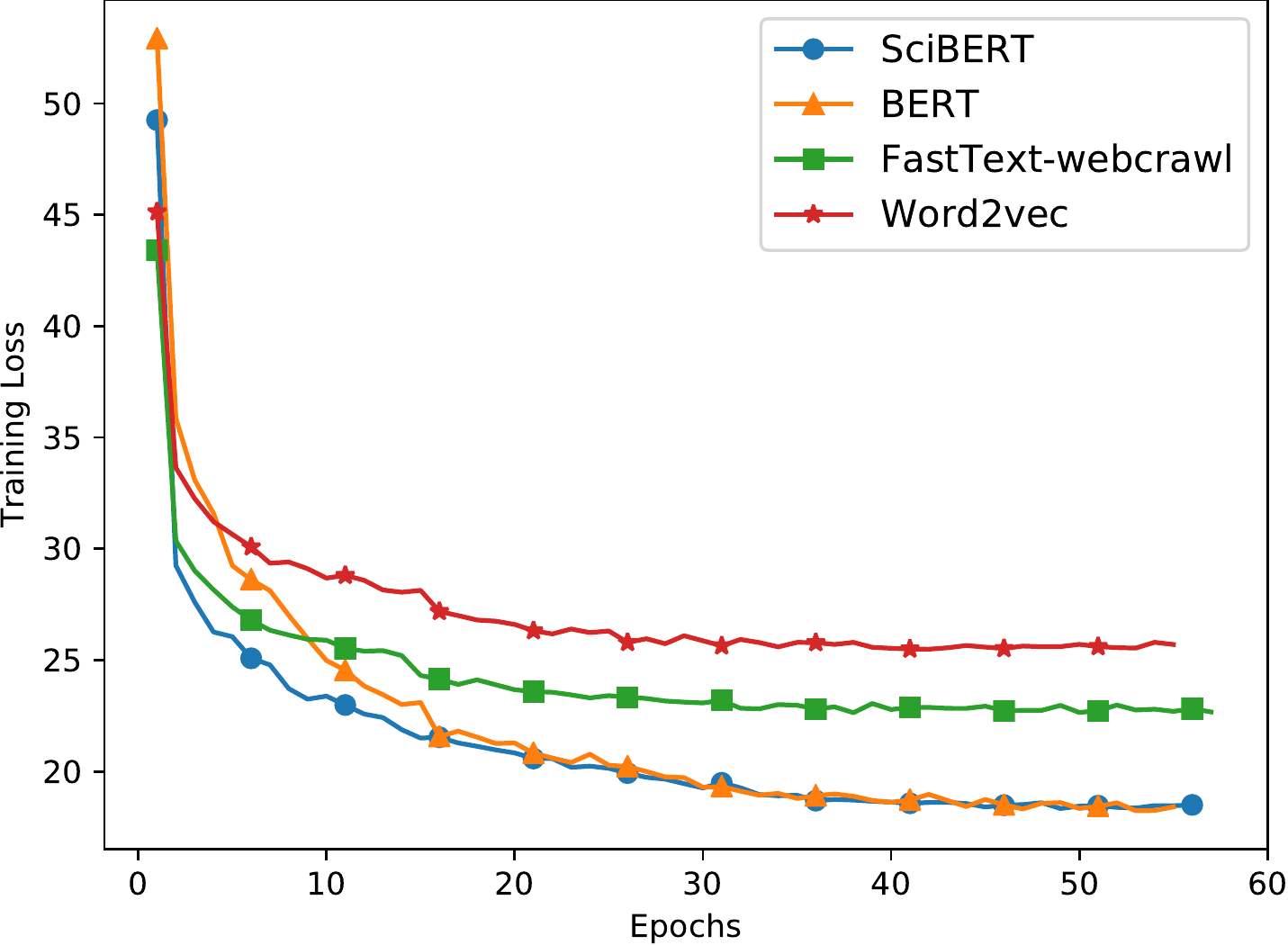}
    \caption{Training loss curve of BLSTM-CRF using BERT and SciBERT vs fixed embeddings for Inspec.}
    \label{fig:train-loss}
\end{figure}

\begin{figure}[htbp]
    \centering
    \includegraphics[width=0.35\textwidth]{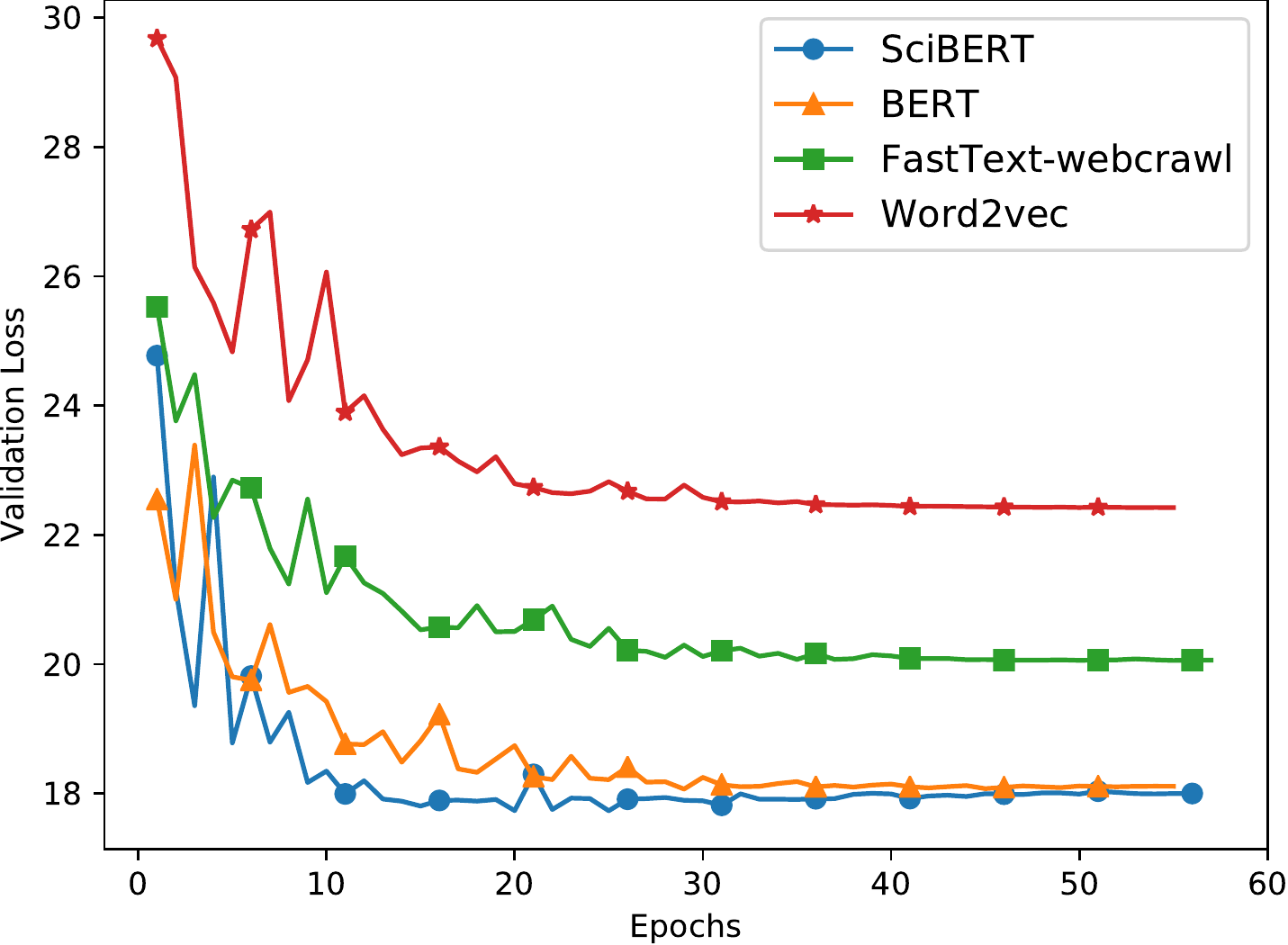}
    \caption{Validation loss curve of BLSTM-CRF using BERT and SciBERT vs fixed embeddings for Inspec.}
    \label{fig:val-loss}
\end{figure}

We also compare the training (Figure \ref{fig:train-loss}) and validation loss (Figure \ref{fig:val-loss}) of BiLSTM-CRF models on four embeddings: SciBERT, BERT, FastText, and Word2Vec. These figures show that the loss values have reduced faster for contextual models than fixed embeddings. Also, SciBERT's validation loss converged faster than BERT. This is consistent with some of the findings in transfer learning literature \cite{goldberg2017neural}, which has shown that these pre-trained contextual models with rich linguistic information can easily adapt to new domains.

\subsection{Baseline comparisons} 
Lastly, we compare our best performing model (BiLSTM-CRF with SciBERT embeddings) against  four baseline methods: SGRank, SingleRank, TextRank, and KEA. Table \ref{tab:sota} presents the results. As expected, our model significantly outperforms all the baseline methods for all three datasets. Of the four baseline methods evaluated here, SGRank  achieved the best results. This observation holds true irrespective of the embeddings used. For example, on Inpsec data, the worst performing BiLSTM-CRF model (Glove at 0.457) is still significantly better than the best performing baseline model (SGRank at 0.271). 

For SemEval-2017, the best reported F1 score is 0.55 \cite{ammar2017ai2}. Ammar et al's work makes use of keyphrase type information in their modeling (if it is a task, material, or process) and they also employs some task specific gazetteers and therefore not directly comparable to our results. Likewise, to our knowledge, the best reported F1 score for SemEval-2010 is 0.29 \cite{mahata2018key2vec} but those models use the entire articles. Though these numbers are not directly comparable to our work, we reported them here to provide a complete context. For Inspec, Mahata et al. \cite{mahata2018key2vec} reported an F1 score of 0.52 which is significantly outperformed by our best model at 0.59.  
%SGRank \cite{danesh2015sgrank}, SingleRank \cite{wan2008single}, TextRank \cite{mihalcea2004textrank}, and KEA \cite{witten2005kea}. 

\begin{table}[htbp]
    \centering
    \scalebox{1}{
    \begin{tabular}{|l|r|r|r|}
         \hline
         & Inspec & SE-2010 & SE-2017 \\
         \hline
         SGRank & 0.271 & 0.229 & 0.211 \\
         SingleRank & 0.123 & 0.142 & 0.155 \\
         TextRank & 0.122 & 0.147 & 0.157 \\
         KEA & 0.137 & 0.202 & 0.129 \\
         BiLSTM-CRF & \textbf{0.593} & \textbf{0.357} & \textbf{0.521} \\
         \hline
    \end{tabular}}
    \caption{\small{Comparison with baseline methods (F1-score)}}
    \label{tab:sota}
\end{table}

\section{Case study: Attention Analysis} 
\label{sec:discussion}

Attention analysis is used to understand if neural network-internal attention mechanisms provide any insight into the linguistic properties learned by the models. We present a case study of attention analysis for keyphrase extraction on a randomly chosen abstract from SemEval2017.

\begin{table}[htbp]
\centering
\scalebox{0.85}{
\begin{tabular}{|p{4.5cm} p{4.5cm}|} 
\hline
SciBERT & BERT \\ 
\hline
An \textcolor{green}{object-oriented version} of \textcolor{green}{SIMLIB} -LRB- a simple simulation package -RRB- This paper introduces an \textcolor{green}{object-oriented version} of \textcolor{green}{SIMLIB} -LRB- an easy-to-understand \textcolor{green}{discrete-event simulation package} -RRB- . The \textcolor{red}{object-oriented version} is preferable to the original procedural language versions of \textcolor{green}{SIMLIB} in that it is easier to understand and \textcolor{red}{teach simulation} from an object point of view . A single-server queue simulation is demonstrated using the object-oriented SIMLIB & An \textcolor{green}{object-oriented version} of \textcolor{red}{SIMLIB} -LRB- a simple simulation package -RRB- This paper introduces an \textcolor{red}{object-oriented version} of \textcolor{red}{SIMLIB} -LRB- an easy-to-understand \textcolor{green}{discrete-event simulation package} -RRB- . The \textcolor{red}{object-oriented version} is preferable to the original procedural language versions of \textcolor{red}{SIMLIB} in that it is easier to understand and \textcolor{red}{teach simulation} from an object point of view . A single-server queue simulation is demonstrated using the object-oriented SIMLIB \\
\hline 
\end{tabular}}
\caption{SciBERT vs BERT: keyphrase identification}
\label{tab:scibert_vs_bert_table}
\end{table}

\begin{figure}[htbp]
    \centering
    \includegraphics[width=0.35\textwidth]{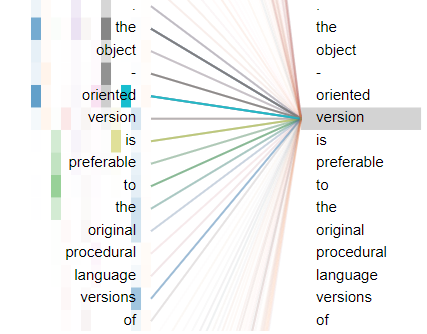}
    \caption{Inadvertent similar token attention in SciBERT}
    \label{fig:layer5_analysis}
\end{figure}

Table \ref{tab:scibert_vs_bert_table} presents the classification results on this abstract from the BERT and SciBERT models;  true positives are marked in green and false negatives in red. Using BertViz \cite{vig2019transformervis} we analyzed the aggregated attention of all 12 layers of both models. We observed that keyphrase tokens ($k_B$ and $k_I$) typically tend to pay most attention towards other keyphrase tokens. Contrarily, non-keyphrase tokens ($k_O$) usually pay uniform attention to their surrounding tokens. We found that both BERT and SciBERT exhibit similar attention patterns in the initial and final layers but they vary significantly in the middle layers. For example, Figure \ref{fig:layer5_analysis} compares the attention patterns in the fifth layer of both models. In SciBERT, the token `object' is very strongly linked to other tokens from its keyphrase but the attentions are comparably weaker for BERT. 

We also observed that keyphrase tokens paid strong attention to similar tokens from other keyphrases. For example, as shown in Figure \ref{fig:layer5_analysis}, the token `version' from `object-oriented version' pays strong attention to `versions' from `procedural language versions'. This is a possible reason for both models failing to identify the third mention of `object-oriented version' in the abstract as a keyphrase. We observed similar patterns in many other documents through our attention analysis. In  future work, we plan to quantify this analysis over multiple documents.

\begin{figure}[htbp]
    \centering
    \includegraphics[width=0.45\textwidth]{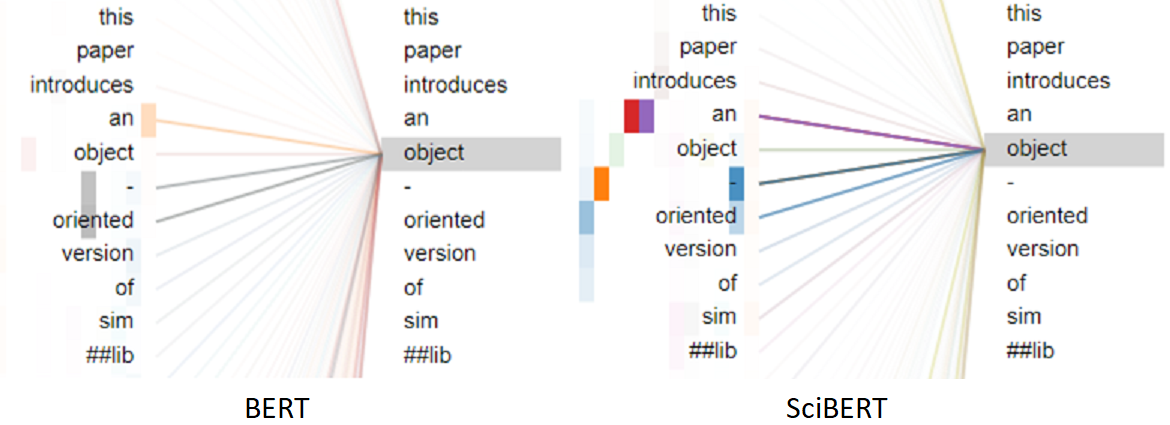}
    \caption{Attention Comparison in Middle Layers}
    \label{fig:layer4_analysis}
\end{figure}

%We also find that SciBERT can detect similar keyphrases correctly, as they pay attention to each other. As a corollary, we observe that SciBERT model isn't able to detect a keyphrase when a token is used in two different contexts in close proximity as the model pays attention to this similar yet different token. This can be visualized in Fig \ref{fig:layer5_analysis} which is the self-attention visualization for Layer 5 of SciBERT. Here the keyphrase token 'version' pays attention to the non-keyphrase token 'versions', and the 'object-oriented version' is not detected as a keyphrase. 

\section{Conclusions}
In this paper, we formulate keyphrase extraction as a sequence labeling task solved using BiLSTM-CRFs, where the underlying words are represented using various contextualized embedding models. Through our experimental work, we quantify the benefits of this architecture over direct fine tuning of the embedding models. We also demonstrate how contextual embeddings significantly outperform their fixed counterparts in keyphrase extraction, with BERT based models performing the best. We also performed attention analysis on a sample scientific abstract to build an intuitive understanding of the working of the self-attention layers of BERT and SciBERT.  

Our approach only deals with the problem of keyphrase extraction but not generation. In the future, we plan to use some of the findings from this paper to help build keyphrase generation models. It would also be beneficial to look at the working of self-attention layers with greater detail and study the fine-tuning capabilities of the contextual embedding models on bigger datasets for the task of keyphrase extraction. We also expect some of the findings in this paper could prove useful to other NLP problems like document summarization.

\fontsize{9.0pt}{10.0pt}
\bibliography{references.bib}
\bibliographystyle{aaai}

\end{document}